\documentclass[letterpaper, 10 pt, conference]{ieeeconf}  % Comment this line out if you need a4paper

\IEEEoverridecommandlockouts                              % This command is only needed if 
\usepackage{graphics} % for pdf, bitmapped graphics files
\usepackage{epsfig} % for postscript graphics files
\title{\LARGE \bf DoShiCo challenge: \\
Domain Shift in Control prediction}

\author{Klaas Kelchtermans$^{*}$ and Tinne Tuytelaars$^{*}$% <-this % stops a space
\thanks{*The authors are with KU Leuven, ESAT-PSI, imec, Belgium.
firstname.lastname@esat.kuleuven.be}% <-this % stops a space
}

\begin{document}

\maketitle
\thispagestyle{empty}
\pagestyle{empty}

%%%%%%%%%%%%%%%%%%%%%%%%%%%%%%%%%%%%%%%%%%%%%%%%%%%%%%%%%%%%%%%%%%%%%%%%%%%%%%%%
\begin{abstract}

%In this work we present the {\em DoShiCo challenge}. It combines two main difficulties: i) training deep control prediction networks end-to-end, and ii) shifting from a simulated to the real-world environment.
Training deep neural network policies end-to-end for real-world applications so far requires big demonstration datasets in the real world or big sets consisting of a large variety of realistic and closely related 3D CAD models. These real or virtual data should, moreover, have very similar characteristics to the conditions expected at test time. These stringent requirements and the time consuming data collection processes that they entail, are currently the most important impediment that keeps deep reinforcement learning from being deployed in real-world applications.  
Therefore, in this work we advocate an alternative approach, where instead of avoiding any domain shift by carefully selecting the training data, the goal is to learn a policy that can cope with it. To this end, we propose the DoShiCo challenge: to train a model in very basic synthetic environments, far from realistic, in a way that it can be applied in more realistic environments as well as take the control decisions on real-world data.
In particular, we focus on the task of collision avoidance for drones.
We created a set of simulated environments that can be used as benchmark and implemented a baseline method, exploiting depth prediction as an auxiliary task to help overcome the domain shift. Even though the policy is trained in very basic environments, it can learn to fly without collisions in a very different realistic simulated environment.
Of course several benchmarks for reinforcement learning already exist - but they never include a large domain shift. On the other hand, several benchmarks in computer vision focus on the domain shift, but they take the form of a static datasets instead of simulated environments. In this work we claim that it is crucial to take the two challenges together in one benchmark.
\end{abstract}
%%%%%%%%%%%%%%%%%%%%%%%%%%%%%%%%%%%%%%%%%%%%%%%%%%%%%%%%%%%%%%%%%%%%%%%%%%%%%%%%
\begin{figure*}[t]
	\centering
    \includegraphics[scale=0.85]{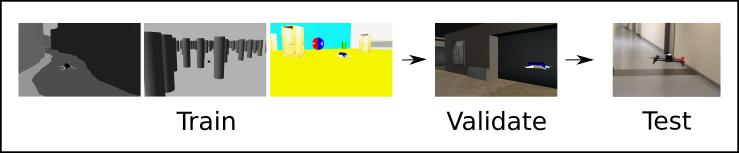}
	\caption{DoShiCo Challenge: train a neural control network end-to-end on three sets of very basic simulated environments (train in "Canyon", "Forest" and "Sandbox"), so that it can fly in a more realistic environment (validate in "ESAT") as well as take control decisions on real-world data (test on the "Almost-Collision dataset") }
    \label{frontpage}
\end{figure*}

\section{Introduction}

% History RL --> time to take step to real world
Reinforcement Learning (RL) is gaining more and more interest due to the strong representational power of deep neural networks (DNNs) ~\cite{Michels2005HighLearning,Abbeel2006AnFlight,Levine2016End-to-EndPolicies,Mnih2013PlayingLearning,Mnih2016AsynchronousLearning}. Where RL used to be a field of low dimensional discrete state-spaces to learn policies succeeding at basic games like tic tac toe \cite{Michie1963ExperimentsParameters},
DNN policies have shown to perform more and more complex tasks on high dimensional \cite{Mnih2013PlayingLearning} and even partially observable state-spaces like first-person views \cite{Wierstra2010RecurrentGradients} and autonomous driving \cite{Udacity2018OpenCar} . 
It is however remarkable that the number of success stories in the form of real-world applications remains low (but see~\cite{Sadeghi2017CAD2RL:Image,Kahn2018Self-supervisedNavigation} for notable exceptions).
In this work we want to take an important step towards the use of deep reinforcement learning (DRL) algorithms in real-world applications by explaining current impediments as well as defining a novel benchmark to stimulate research in this direction.

% Unfeasibility of training directly in the real world: fatal crashes
Using RL algorithms to train DNN policies in the real world is highly impractical due to several reasons. The first and most obvious one is the fact that most algorithms are to be trained on-policy as they are based on trial-and-error. For tasks like autonomous navigation, this means that a suboptimal policy will be steering the robot or drone, leading to possibly fatal crashes and rather unhappy researchers.

% Unfeasibility of training directly in the real world: big amount of experiences required
A second reason is the large amount of training experiences that is required. Even if your algorithm could train from an offline dataset,
%, e.g. using imitation learning from demonstrations, 
it would require numerous hours of demonstrating the same task over and over again. In~\cite{Gandhi2017LearningCrashing}, it is shown that collecting a dataset of 11,500 crashes of a drone allows training a policy to avoid obstacles, though it is clear that this strategy is not an option for many applications.

% Other reasons and conclusion to simulation
Moreover, there are other reasons, like the exploitation of parallelism, the possibility to reset the robot to a certain state and facilitating the reproduction of results, that make training DNN policies most feasible in simulated environments.

% Introduction to domain shift and alternative solutions
This, however, introduces a novel problem when testing the policy in the real world. Namely, the simulation and the real world will never look or act exactly alike. This is referred to as a {\em domain shift} between the input at training and test time. There have been several attempts to deal with this issue, though none of them seems to head in a promising direction: researchers have tried to make the simulated environment photo-realistic \cite{Richter2016PlayingGames}, or to introduce such a large variance over simulated environments in order to generalize to the real world \cite{Sadeghi2017CAD2RL:Image}, or lastly to augment the training data with domain shift techniques so it looks more and more like the real world \cite{Yoo2017AutonomousAdaptation}. 
These techniques seem to come with a lot of overhead if we want to apply them to more specific tasks, e.g. autonomous surveillance of a particular building site or autonomous inspection of a windmill. Instead of putting effort in data collection or in building simulated environments for each specific task, i.e. trying to minimize the discrepancy between train and test conditions, it is worth exploring alternative strategies, that can effectively cope with this domain shift.

% Computer vision benchmarks focussing domain shift never incorporate control prediction
In computer vision, several benchmarks exist that focus on the domain shift for different tasks like image segmentation and depth prediction \cite{Geiger2012AreSuite,Cordts2016TheUnderstanding,Peng2017VisDA:Challenge}. We argue however that for the setting of control prediction or policy training, it is crucial to evaluate and preferably train in an online fashion. This is due to the non-i.i.d. (independent and identically distributed) nature of sequential decision processes. If a mistake is made at a certain point in time, a compound error over time will lead to very different outcomes. In that sense, evaluating solely on a static dataset can never fully represent the on-policy performance. In this work we propose a challenge in the form of train and test environments instead of static data. 

% RL benchmarks focussing on POMDP never incorporate domain shift
On the other hand within the RL community, benchmarks exist that compare different DRL training algorithms like THOR \cite{Roozbeh2017THORChallenge}, VIZDOOM \cite{Kempka2016ViZDoom:Learning}, OpenAI \cite{Brockman2016OpenAIGym}, CARLA \cite{Dosovitskiy2017CARLA:Simulator}, TORCS \cite{Wymann2015TORCS:Simulator}, Udacity \cite{Udacity2018OpenCar}, etc. However, they never incorporate a large domain shift. To the knowledge of the authors, there does not exist a benchmark that combines the two challenges: solving a Partially-Observable Markov Decision Process (POMDP) by training a DNN policy end-to-end, together with the domain adaptation to the real-world. By introducing this challenge of training DNN policies end-to-end, we believe we can boost this research field, taking DRL algorithms to the next step towards real-world applications.

% General explanation of the goal
This requires policies that can generalize to previously unseen conditions, perform well over a wide action range while at the same time being insensitive to irrelevant differences, all learned in an end-to-end fashion. In the case of obstacle avoidance, this could be realized by exploiting visual cues such as relative pose and depth and by learning invariance to color or texture. Note that the question whether this policy is trained in a supervised fashion or in a reinforced fashion remains open, i.e. the benchmark can be used for both.

% Contribution
Building on the above observations, we make two contributions. First, we propose a domain shift challenge for control prediction (DoShiCo) that can serve as a benchmark for comparing different training strategies. Given three sets of very basic simulated environments, instances of which can be generated randomly during training, the goal is to train a DNN policy end-to-end for the task of collision avoidance so that it can generalize to previously unseen and more realistic conditions, as captured by our synthetic yet more realistic validation environment and real-world test data (see Figure \ref{frontpage}). To avoid the issue with online control in a real-world setting, a classification task using a dataset of 'almost collisions' is provided as a proxy for flying in the real world. 

Second, we evaluate a two baselines in this setting, building on a MobileNet~\cite{Howard2017MobileNets:Applications} pretrained on ImageNet~\cite{Russakovsky2015ImageNetChallenge}. In an attempt to train the control end-to-end without losing the robustness to varying imaging conditions and without overfitting to the basic simulated environments, we demonstrate the benefit of an auxiliary task \cite{Jaderberg2017ReinforcementTasks}.
The use of auxiliary tasks helps the extracted features to focus on the information that is relevant for the task which makes the learning less prone to fitting toward irrelevant features only visible in the training environments.
In particular, we demonstrate how auxiliary depth prediction can reduce the impact of the domain shift and we show that our model succeeds in flying in the more realistic simulated validation world although it was trained solely on a mix of very basic environments.  %The policy trained could even perform a couple of short real-world flights online, as can be seen in the supplementary material. 

In order to use DoShiCo as a benchmark, we integrated the full setting of ROS, Gazebo and Tensorflow in a Docker and Singularity image. 
All the code, 3D environments and the best trained checkpoints of the model are publicly available to ensure reproducibility of the results\footnote{kkelchte.github.io/doshico}.

The remainder of this paper is organized as follows. First, in section~\ref{sec:background}, we start with a short background, defining monocular obstacle avoidance in an RL setting. Second, in section~\ref{sec:related}, we describe the related work. Next, we give the details of the DoShiCo challenge (Section~\ref{sec:challenge}) and details on our baseline models (Section~\ref{sec:model}). In Section~\ref{sec:expres}, the experimental results are discussed. Section~\ref{sec:conclu} concludes the paper together with a discussion.

\section{Background}
\label{sec:background}

% In the case of a fully RL problem, % TT: wat bedoel je met 'fully' ? 
% KK: als tegenstelling tot dynamic programming (environment dynamics are known) of
% multi-armed bandits (reward = return)
In the general RL setting,
an agent interacts with the environment according to a policy, $\pi$, which maps the currently observed state $s_t$ to an action $a_t = \pi(s_t)$. The environment brings the agent to a next state $s_{t+1}$ according to the dynamics of the environment $T(s_{t+1}|s_t,a_t)$ with a corresponding reward $r_{t+1}$. 
The goal is then to find good parameters for this policy, maximizing the cumulative reward in the longterm~\cite{Sutton2017ReinforcementIntroduction}. 

Most of the RL algorithms are based on the Markov property. In this case, the information provided in the state should be enough to select an optimal action. 
Such problems are referred to as Markov Decision Processes (MDPs). In most applications, however, this assumption does not hold. In those cases, the process is seen as an MDP but the agents observation is not the full state. These problems are referred to as partially observable MDPs (POMDP). % this means that the agent needs to estimate its current state given a certain observation.

As mentioned in the intro, in state-of-the-art RL methods, states have shifted from low dimensional game specific variables to high dimensional raw images, and further to partially observable states for example in first-person views.

% TT: Dit hoort eigenlijk niet bij 'background' - moet later nog ergens herhaald worden voor lezers die de background section overslaan.
% KK: toegevoegd aan intro doshico challenge.
In the DoShiCo challenge the observed state is represented by the current view of a drone. The action is the applied yaw turn while flying at a fixed speed. 
% Our baseline method is learned with imitation learning in which case rollouts of the game is provided by an expert demonstrating the desired behavior
% Learning the policies becomes minimizing the imitation loss as difference in prediction

%%%%%%%%%%%%%%%%%%%%%%%%%%%%%%%%%%%%%%%%%%%%%%%%%%%%%%%%%%%%%%%%%%%%%%%%%%%%%%%%
\section{Related Work}
\label{sec:related}

\subsection*{Learning to control} When it comes to autonomous navigation directly from high dimensional camera input, a first family of solutions relies on geometric techniques for simultaneously mapping and localizing the agent (SLAM). Besides the extra computational power, these algorithms often suffer from a lack of features to track \cite{Engel2016DirectOdometry}. In a similar spirit yet more robust, Gupta et al. \cite{Gupta2017CognitiveNavigation} have recently demonstrated how a joint neural architecture, called a CMP (Cognitive Mapping and Planning), can effectively learn to map and plan jointly trained in an end-to-end fashion.

% Using intermediate representations
More related to our work, yet avoiding the end-to-end complexity and bypassing the domain shift issue, there are some works that rely on depth estimation as an intermediate step. Michels et al.~\cite{Michels2005HighLearning} predict depth and then train a controller in simulation with reinforcement learning. 
In \cite{Chakravarty2017CNN-basedQuadrotor} depth is estimated from single images using a CNN, as in~\cite{Eigen2014DepthNetwork}, and then used to avoid obstacles with a behavior arbitration algorithm~\cite{Althaus2002BehaviourEnvironments}. For the task of obstacle avoidance this might be a valid solution. There are however different and more complex tasks for which depth estimation does not provide enough information. In those settings end-to-end training is required in order to extract higher level features. 

% pioneers of vision-based end-to-end control
Moreover, Levine et al. \cite{Levine2016End-to-EndPolicies} demonstrated that training end-to-end results in more stable and efficient learning. 
%In fact, early work already demonstrated the promising path of collision avoidance control directly from RGB input. 
Pomerleau \cite{Pomerleau1991RapidlyNavigation} successfully trained a single layer network end-to-end with apprenticeship learning for the task of road following. Lecun et al.~\cite{Lecun2005Off-RoadLearning} trained a CNN to predict the steering angle of a small car based on stereo input from an offline dataset.  
In more recent work, Ross et al.~\cite{Ross2013LearningEnvironments} train an SVM (Support Vector Machine) iteratively with imitation learning in a forest. 
Giusti et al.~\cite{Giusti2016ARobots} train a deep neural network end-to-end for following trails in a forest from a large offline dataset gathered manually.

These are all examples demonstrating the promising path of end-to-end image-based control prediction networks.

% domain shift
\subsection*{Dealing with the virtual-real domain shift} All control models previously mentioned are trained in an environment that is close to the test environment. However, as indicated above, for many real-world applications this is unfeasible. There have been some preliminary attempts to cope for instance with different weather conditions \cite{Daftry2016LearningControl, Dosovitskiy2017CARLA:Simulator}. In the recent work \cite{Li2018LearningGeneralization} of Li et al., they succeed at generalizing over different game parameters in a Cart-Pole game (cart mass and pole length). The step from simulation to the real world entails however a much bigger domain shift. In computer vision, several methods for domain adaptation have been proposed (see~\cite{Csurka2017DomainSurvey} for a recent survey), but mostly in rather artificial setups and, to the best of our knowledge, never in the context of control prediction networks. 

% The work coming closest to a solution to this problem is the CAD2RL setup \cite{Sadeghi2017CAD2RL:Image}. Here the deep neural policy is trained with reinforcement learning in a large variety of realistic, though not photorealistic, 3D CAD models of hallways. Due to the large visual variance and realistic models in simulation the policy can generalize to real hallways. However, using a large variety of realistic 3D models to close the gap between simulation and the real-world, seems an inefficient way to deal with the issue. Besides, it is often impossible to obtain such a large variety of realistic 3D models for a specific real-world application. 

Finally, the idea of using auxiliary tasks for our baseline model stems from Multi-Task Learning (MTL). In computer vision, MTL has been demonstrated to improve the performance of one task by sharing the network with other tasks -- for instance object detection together with classification and segmentation \cite{Caruana1997MultitaskSeptember1997}. Mirowski et al. \cite{Mirowski2017LearningEnvironments} demonstrate the use of depth prediction as an auxiliary task in order to learn a deep neural agent to navigate and localize itself in a simulated maze. They demonstrate a positive impact on data efficiency and performance, however they did not research the domain shift.

% General critic: how do you know that auxiliary task isn't just improving the overall performance of solving the POMDP
% and do you think that it is influencing the domain shift:
% My response: does it matter? 
% If you have an RL algorithm that finds a policy that performs very well in these basic environments it is hard to improve, 
% while if the policy can also predict some depth at the same time, this policy will be more robust to domain shifts.

%%%%%%%%%%%%%%%%%%%%%%%%%%%%%%%%%%%%%%%%%%%%%%%%%%%%%%%%%%%%%%%%%%%%%%%%%%%%%%%%
\section{The DoShiCo Challenge}
\label{sec:challenge}
The DoShiCo challenge contains of training a DNN policy end-to-end for the task of obstacle avoidance with drones. 
Based on the high-dimensional RGB input from the current view of the drone, the policy needs to predict the action, in this case the applied yaw turn while flying at a fixed speed. 
The policy is trained in very basic simulated environments, chosen specifically to train features relevant for obstacle avoidance. After this, the policy is evaluated \emph{online} in a very different and more realistic simulated environment and \emph{offline} on a small dataset of real-world videos.
%The DoShiCo challenge contains of training in a mix of 3 types of environments: Canyons, Forests and Sandboxes. The performance of the policy is be evaluated online and on-policy in a more realistic ESAT environment. The difference between the very basic training and more realistic evaluation environment entails already a dummy domain shift. Finally the performance is tested on a offline classification task with real-world data from the almost-collision dataset.

The DNN agent should be initialized from scratch or with imagenet-pretrained weights. Pretraining the neural network with intermediate representations, like depth maps, separates the two challenges and is therefore not allowed. The policy should be trained end-to-end. The challenge does not restrict the training time or the architecture of the network.

\subsection*{A Mix of Basic Training Environments: Canyons, Forests and Sandboxes}

\begin{figure}[tbp]
	\centering
	\includegraphics[scale=0.65]{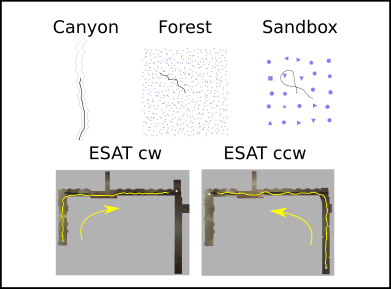}
	\caption{Top-down view of the training (top) and validation (bottom) environments with an example trajectory.}
    \label{expsenvs}
\end{figure}

Snapshots of the type of environments are shown in Figure \ref{frontpage}. Top-down views can be seen in Figure \ref{expsenvs}.
One type of training environments is a canyon that bends randomly, inspired by~\cite{Kahn2017PLATO:Optimization}. The goal is to fly more than 45m through the canyon at a constant speed of 1.3m/s without colliding with the walls. 
From the same work, we copied the idea of a forest with cylinders placed on random spots in which the goal is to cross 45m without collision.
The third environment is called the Sandbox. It is a box of 20x20m with walls in varying colors and a number of different objects spread around. We picked 13 basic objects with different shapes found on the Gazebo model server \cite{OpenSourceRoboticsFoundation2018Gazebo}. In the sandbox the agent needs to get further than 7m from the starting position without collision.
In each environment the agent is spawned at location $(0,0)$. The environments are made on the fly. 

The three training environments intuitively help to learn a different type of behavior.
In the canyon, the agent learns to focus on perspective lines relevant for wall- and corridor-following. In the forests, the agent learns to avoid moving vertical lines relevant for general obstacle avoidance. The sandboxes ensure a healthy invariance towards a variety of shapes and visual features.

\subsection*{More realistic validation environment: ESAT}
The network is evaluated \emph{online} in the ESAT environment. We build a lookalike model of our corridor at the electrical engineering department ESAT. It is important to test online in order to know how well the agent has solved the POMDP. The policy is tested through the ESAT corridor in 2 directions. Successful example flights are depicted in the top-down view of Figure \ref{expsenvs}.

\subsection*{Real-world test data: Almost-Collision dataset}
Collision avoidance on a real drone performed by different policies is hard to compare correctly. Real-world experiments are influenced by many external factors such as battery state, propeller state, on board electronics, etc..
A drone might crash in one test, deteriorating all consecutive flights. 

Alternatively using an imitation loss from a demonstration flight by a pilot is biased towards the specific flying behavior of the pilot which is unfair for comparing one policy to another. 
%Different policies might prefer flying more on the left side or the right side of a corridor while both avoiding collision successfully. 

In order to compare collision avoidance in a fair and quantitative way on real-world data, we made a small dataset containing sequences of images of situations in which only one control is suitable: straight, left or right turn.  We make sure that collision is very nearby in all trajectories, without actually crashing. This conveniently resulted in the Almost-Collision dataset.

\begin{figure}[tbp]
	\centering
	\includegraphics[scale=0.7]{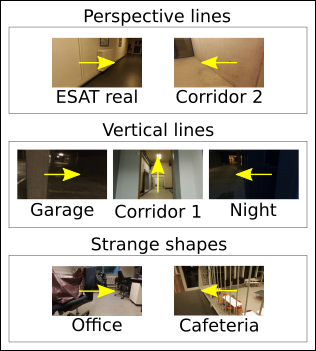}
	\caption{Snapshots from the seven different locations of the Almost-Collision dataset. The yellow arrow indicates the target direction. Different trajectories have different visual cues to indicate the approaching collision.}
    \label{almost_collision}
\end{figure}

We recorded data on seven different locations that differ a lot in visible features. Snapshots are shown in Figure \ref{almost_collision}. The trajectories are tagged with different visible cues. These cues are a type of feature specific to this type of collision: perspective lines, vertical lines and strange shapes. 
%The cues are learned in the corresponding simulated environments: Canyon, Forest and Sandbox.

The image sequences are around 3 to 5 seconds at 20fps. They are labeled with the control required to avoid collision. Over the seven locations a total of 25 trajectories are collected with an equal amount of left and right target controls with the exception of one trajectory with straight as target control. The total size is around 1600 frames. For classification, the predicted angular velocity in yaw is discretized with thresholds $\pm0.3$ for left, straight and right.

\subsection*{Evaluation}

In order to evaluate a model on the DoShiCo challenge it is important to see first how well the model has solved the POMDP in the three training environments online. The second evaluation is also online in the ESAT environment for which it needs to cope with a primal domain shift. This second criteria is the most important one as it combines the difficulty of a domain shift together with the on-policy POMDP setting. The final criteria is an indication for the use of this agent in a real-world obstacle avoidance scenario. 

During our experiments we encountered very high variance in the online results of policies trained with different seeding. This made it hard to compare different setups as well as hyperparameters. 
%Though the variance seems to be a common problem in the field of training deep neural policies, it is not always openly reported. 
Jaderberg et al. \cite{Jaderberg2017ReinforcementTasks} compare performances of the top 3 policies picked from 50 policies trained with different hyperparameters. We found that plotting the performance expressed as a percentage of the ranked population allows a proper comparison.

%%%%%%%%%%%%%%%%%%%%%%%%%%%%%%%%%%%%%%%%%%%%%%%%%%%%%%%%%%%%%%%%%%%%%%%%%%%%%%%%
\section{Baseline Model}
\label{sec:model}
Here, we describe our training strategy as well as the architectures of the models used in our experiments. Note that the DoShiCo challenge does not provide any restrictions on how the policy is trained. 

\subsection*{Training strategy}
Our baseline model is trained in a straightforward imitation learning way, namely behavioral cloning. This means that the data is collected by flying a number of times through the simulated training environments. This was automated with a heuristic based on groundtruth depth provided by the simulated environment. The policy is trained on this data in a supervised fashion, minimizing the difference between the estimated control and the control applied by the heuristic. The heuristic collects data of 100 flights in each of the three types of training environments. 

\subsection*{Architectures: NAUX \& AUXD}
\begin{figure}[t]
	\centering
% 	\framebox{\parbox{3in}{\includegraphics[scale=1.0]{frontpage.png}}}
	\includegraphics[scale=0.5]{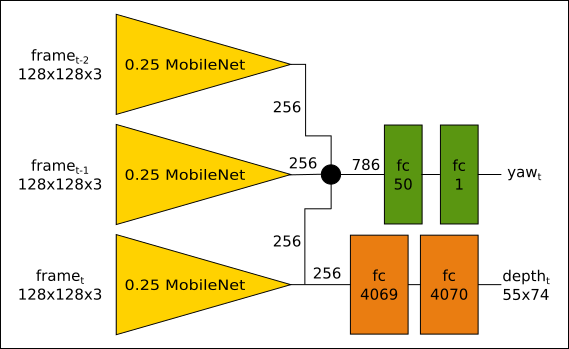}
	\caption{The architecture of the policy. The figure is best seen in color. The yellow parts are 0.25-MobileNets that share weights over three consecutive frames. The circle represents the concatenation of the extracted features. The green part is the control prediction that consists of two fully connected layers. The orange part represents the auxiliary depth prediction layers.}
    \label{architecture}
\end{figure}

The architecture of the baseline model is shown in Figure~\ref{architecture}. The base network is called NAUX (for "No Auxiliary task") and contains a feature extracting part (yellow) and two fully-connected layers for control (green). In order to give the neural network a sense of time, the network takes three consecutive frames as input. Each frame is fed to a feature extracting part with shared weights, with the architecture of mobilenet-0.25 \cite{Howard2017MobileNets:Applications}. The weights of the mobilenet-0.25 are initialized from a model pretrained on Imagenet \cite{Russakovsky2015ImageNetChallenge}. The features are concatenated (black circle) and fed to the control prediction part (green). The control part has a fully connected hidden layer with 50 nodes with ReLu activation and an output layer with no activation function. 
% we experienced that the hyperbolic tangent function restricts the steering too close to 0. 
The control output is a continuous value of the angular velocity in yaw. 
% By steering the drone over the yaw angle and not strafing with the roll, we ensure that in case of low drift the camera is always looking in the flight direction. 

The network with auxiliary depth prediction is further referred to as AUXD. It is build on top of NAUX. The extracted features of the last frame are fed to two fully connected layers that predict a depth frame of 55x74, based on the work of Eigen et al. \cite{Eigen2014DepthNetwork} (see Figure~\ref{architecture}).

\section{Experimental Results}
\label{sec:expres}

\begin{table}[tbp]
\caption{Online Performance in Simulation}
\begin{center}
\begin{tabular}{|r||c|c|c|c|c|c|}
\hline
& \multicolumn{6}{c|}{Average distance [m]} \\ \hline
& \multicolumn{2}{c|}{TOP 5} & \multicolumn{2}{c|}{TOP 3} & \multicolumn{2}{c|}{TOP 1} \\ \hline
& NAUX & AUXD & NAUX & AUXD & NAUX & AUXD \\ \hline \hline
Canyon & 43.96 & 38.41 & 42.33 & 43.08 & 38.05 & 41.79 \\ \hline
Forest & 45.99 & 50.24 & 42.90 & 48.48 & 48.67 & 51.35 \\ \hline
Sandbox & 7.03 & 8.62 & 6.98 & 9.11 & 9.22 & 8.09 \\ \hline \hline
ESAT & 47.03 & 57.63 & 50.08 & 61.66 & 60.25 & 71.69 \\ \hline
\end{tabular}
\end{center}
\label{aux_simulation}
\end{table}
%\subsection*{NAUX versus AUXD: The influence of depth prediction as an auxiliary task.}

\begin{figure}[tbp]
	\centering
    \includegraphics[scale=0.6]{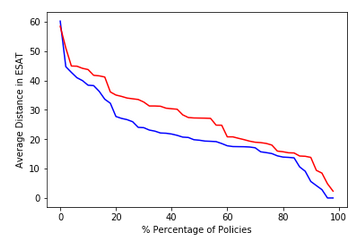}
	\caption{The variance of the on-policy performance as distance [m] traveled in the ESAT validation environment over the percentage of the population of policies. The red and blue lines correspond to the NAUX and AUXD architectures.}
    \label{variance}
\end{figure}

 %Inspired by this work, we plot a similar graph in Figure \ref{variance}. 
 The performance is measured as the average collision-free distance traveled by the policy online over 10 runs in the ESAT environment. The population over 50 policies is visible in Figure \ref{variance}. Please note that the evaluation on the more realistic ESAT environment entails a primal domain shift which augments the variance over the different trained policies. 

The blue line is the ranked performance of the NAUX networks trained without auxiliary depth. The red line defines the AUXD networks trained with depth prediction. It is clearly visible how the use of auxiliary depth improves the general online performance on the validation environment.

Table \ref{aux_simulation} shows the performance averaged over the top 5, top 3 and best policies. The policies are selected based on their average distance traveled online in the ESAT environment. The first three rows are the online performances of both NAUX and AUXD policies tested in new generated environments similar to the training environment. The performances are expressed as average collision-free distance. % and average number of successes out of 10 runs. 

Evaluating the performance on environments similar to the training environments is the common practice in training deep neural control. It is clear how, both with and without auxiliary depth, the policies can already succeed a large number of times. In other words, it has learned to succeed at avoiding collisions in the basic environments. Training on longer and more data with possibly intermediate on-policy iterations or larger networks could improve these numbers further but that is not the goal of this work.

% The last row of the table shows the average performance of the best policies online in the ESAT environment. It is remarkable that the top networks succeed at performing this task. The ESAT validation environment is not only visually very different, also the trajectory is very different. The distance is much longer and the turns of 90 degrees are not present in the Canyon training environments. Moreover the network is trained off-policy from a dataset so has never actually flown before.

As Figure \ref{variance} implied, the auxiliary depth has a consistent positive impact on the validation performance. This positive impact is less present in environments similar to the training environment which confirms the believe that the auxiliary task helps to regularize over a domain shift. 

\begin{table}[t]
\caption{Accuracies on the Almost-Collision dataset [\%] \newline NA $\sim$ NAUX and AD $\sim$ AUXD }
\begin{center}
\begin{tabular}{|r||c|c|c|c|c|c|}
\hline
& \multicolumn{2}{c|}{TOP 5} & \multicolumn{2}{c|}{TOP 3} & \multicolumn{2}{c|}{TOP 1} \\ \hline
& NA & AD & NA & AD & NA & AD \\ \hline 
ESAT real & 35 & \textbf{64} & 28 & \textbf{80} & 27 & \textbf{73} \\ \hline 
Corridor 1 & 40 & \textbf{57}& 38 & \textbf{61}& 40 & \textbf{60} \\ \hline 
Corridor 2 & \textbf{53} & 39& \textbf{51} & 45& \textbf{49} & 21 \\ \hline 
Office & 76 & \textbf{99}& 78 & \textbf{100}& 100 & 100 \\ \hline
Cafeteria & 29 & 25 & 30 & 26 & \textbf{42} & 34 \\ \hline 
Garage & 50 & \textbf{62}& 45 & \textbf{56}& 46 & \textbf{58} \\ \hline 
Night & \textbf{77} & 63& 72 & 71& \textbf{76} & 59 \\ \hline
Avg. Loc. & 51 & \textbf{58}& 49 & \textbf{63}& 54 & 58 \\ \hline \hline
Strange & 46 & \textbf{54}& 45 & \textbf{57}& 52 & \textbf{60}\\ \hline 
Perspective & 46 & 49& 43 & \textbf{53}& 46 & 44\\ \hline 
Vertical & 70 & 72& 68 & \textbf{76}& 66 & \textbf{73}\\ \hline 
Avg. Cue & 54 & 58& 52 & \textbf{62}& 55 & 59\\ \hline 
\end{tabular}
\end{center}
\label{aux_real}
\end{table}

The step to the real world comes with an extra domain shift. The best policies are quantitatively evaluated on the Almost-Collision dataset. The results are shown in table \ref{aux_real}. The top rows show the results per location and with an overall average taken with equal weight for each location. The bottom rows show the results per type of visual cue with an overall average taken with equal weight for each visual cue.
The best accuracy between NAUX and AUXD is put in bold if the difference is significant (greater than 5\%). The improvement is less distinct with auxiliary depth on real world data. Although the average top 5 and top 3 performances indicate a positive trend towards the use of auxiliary depth.

\section{Discussion and conclusion}
\label{sec:conclu}

The ability of neural networks to be trained in simulation and still perform robustly in the real world is a major challenge for applying deep neural control in real-world applications. A lack of benchmarks that combines the online evaluation environment with a domain shift, makes it hard to compare different methods. 

You could argue that the general task of obstacle avoidance is already solved once you can predict proper depth maps over different domains as intermediate features. This is true, however the goal of DoShiCo is not to solve monocular obstacle avoidance. The goal is to find ways of training DRL policies in such way that they generalize over different domains. In many real-world applications it might not be convenient to find informative intermediate representations that can be trained separately. 

As a baseline we propose a model that successfully uses auxiliary depth prediction learned in a behavioral cloning fashion. Our model succeeds at taking the dummy domain shift from the basic mixed environments to the ESAT validation environment. Though the performance on real-world data still has to be improved. %It is even capable of competing with a model trained in a more realistic environment when tested on the Almost-Collision dataset.

As many questions remain unanswered: 1. Is it best to train a model with imitation learning or fully reinforced when looking at such a large domain shift? 2. How can different architectures help to generalize over different domains while still allowing end-to-end training? 3. What is the influence of a discrete versus continuous action-space on the generalization ability? 
Further study will have to demystify the training of image-based DNN policies. 
By proposing the DoShiCo challenge, we want to boost this research field, taking RL a step further towards the real-world.

% The use of temporal features with recurrent neural networks might improve robustness in combination with extra techniques from domain adaptation. So far the model is trained with supervised learning. Techniques from imitation learning and reinforcement learning might as well stabilize the training reducing the variance introduced by the domain shift. 

\section*{Acknowledgements}
This  work  was supported  by the CAMETRON research project of the KU Leuven (GOA).

\addtolength{\textheight}{-12cm}   % This command serves to balance the column lengths

%%%%%%%%%%%%%%%%%%%%%%%%%%%%%%%%%%%%%%%%%%%%%%%%%%%%%%%%%%%%%%%%%%%%%%%%%%%%%%%%

%%%%%%%%%%%%%%%%%%%%%%%%%%%%%%%%%%%%%%%%%%%%%%%%%%%%%%%%%%%%%%%%%%%%%%%%%%%%%%%%

%%%%%%%%%%%%%%%%%%%%%%%%%%%%%%%%%%%%%%%%%%%%%%%%%%%%%%%%%%%%%%%%%%%%%%%%%%%%%%%%
% \section*{Appendix}

% \section*{Acknowledgment}

%%%%%%%%%%%%%%%%%%%%%%%%%%%%%%%%%%%%%%%%%%%%%%%%%%%%%%%%%%%%%%%%%%%%%%%%%%%%%%%%

%\bibliographystyle{IEEEtran}
\bibliographystyle{ieeetr}
\bibliography{Mendeley}

\end{document}